%% file: acl_latex.tex
\pdfoutput=1

\documentclass[11pt]{article}

\usepackage{arydshln}  
\usepackage{graphicx}  
\usepackage{amsmath}   
\usepackage[table]{xcolor}
\usepackage{enumitem}

\usepackage[preprint]{acl}

\usepackage{times}
\usepackage{latexsym}

\usepackage[T1]{fontenc}

\usepackage[utf8]{inputenc}

\usepackage{microtype}

\usepackage{inconsolata}

\usepackage{graphicx}
\usepackage{amsmath}     
\usepackage{amssymb}     
\usepackage{mathtools}   
\usepackage{algorithm, algorithmic}
\usepackage{amsthm}
\graphicspath{ {./figures/} }

\usepackage{setspace}

\usepackage{etoolbox}

\usepackage{adjustbox}
\usepackage{hyperref}

\usepackage{algorithm}
\usepackage{amsmath}
\usepackage{amssymb}

\title{GEMMAS: Graph-based Evaluation Metrics for Multi Agent Systems}

\author{Jisoo Lee\thanks{Equal Contribution.} \\
  Seoul National University \\
  \texttt{sally66890@snu.ac.kr} \And
  Raeyoung Chang\footnotemark[1] \\
  Sogang University \\
  \texttt{icanry@sogang.ac.kr} \And
  Dongwook Kwon\footnotemark[1] \\
  Kwangwoon University \\
  \texttt{dongwook.kwon@kw.ac.kr} \AND
  Harmanpreet Singh\thanks{Corresponding author.} \\
  LG Electronics, Toronto AI Lab \\
  \texttt{harmanpreet.singh@lge.com} \And
  Nikhil Verma\footnotemark[2] \\
  LG Electronics, Toronto AI Lab \\
  \texttt{nikhil.verma@lge.com}}

\begin{document}
\maketitle

\input{body/abstract}
\input{body/1_introduction}
\input{body/2_related_work}
\input{body/3_method}
\input{body/4_results_analysis}
\input{body/5_conclusion}




\bibliographystyle{acl_natbib}
\bibliography{references}

\clearpage
\newpage
\appendix
\input{body/appendix}

\end{document}

%% file: body/abstract.tex
\begin{abstract}

Multi-agent systems built on language models have shown strong performance on collaborative reasoning tasks. 
However, existing evaluations focus only on the correctness of the final output, overlooking how inefficient communication and poor coordination contribute to redundant reasoning and higher computational costs.
We introduce GEMMAS, a graph-based evaluation framework that analyzes the internal collaboration process by modeling agent interactions as a directed acyclic graph. 
To capture collaboration quality, we propose two process-level metrics: Information Diversity Score (IDS) to measure semantic variation in inter-agent messages, and Unnecessary Path Ratio (UPR) to quantify redundant reasoning paths.
We evaluate GEMMAS across five benchmarks and highlight results on GSM8K, where systems with only a 2.1\% difference in accuracy differ by 12.8\% in IDS and 80\% in UPR, revealing substantial variation in internal collaboration.
These findings demonstrate that outcome-only metrics are insufficient for evaluating multi-agent performance and highlight the importance of process-level diagnostics in designing more interpretable and resource-efficient collaborative AI systems.

\end{abstract}

%% file: body/1_introduction.tex
\section{Introduction}
\label{sec:introduction}

\begin{figure*}[htb]  
    \centering
    \includegraphics[width=\linewidth, keepaspectratio]{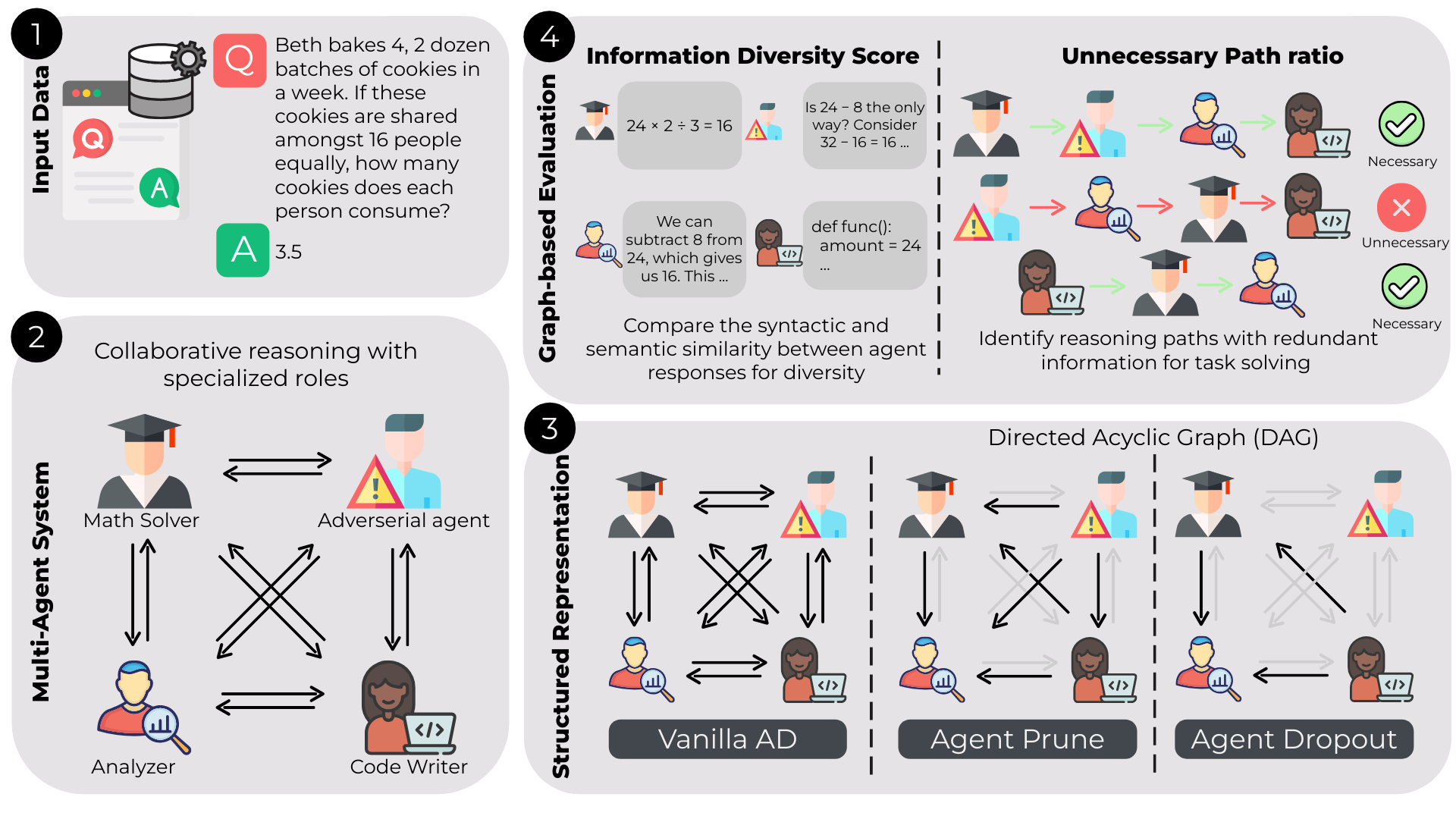}
    \caption{\textbf{Overview of the GEMMAS evaluation framework.} The process begins with input mathematical problems, which are solved collaboratively by a multi-agent system composed of specialized agents. Their interactions are represented as a DAG, capturing both communication flow and reasoning structure. From this DAG, GEMMAS computes structural metrics—Information Diversity Score (IDS) and Unnecessary Path Ratio (UPR)—to evaluate collaboration quality and efficiency beyond final-task accuracy.}
    \label{fig:fig1}
\end{figure*}

Large language models (LLMs) such as GPT-4~\cite{achiam2023gpt}, Llama~\cite{touvron2023llama}, and Qwen~\cite{bai2023qwen} demonstrate emergent reasoning capabilities and achieve state-of-the-art performance across a variety of NLP tasks.
As multi-agent LLM systems become more prevalent~\cite{shen2023hugginggpt,chen2023agentverse}, their evaluation remains narrowly focused on the correctness of the final answer. 
This outcome-centric view overlooks the underlying collaboration dynamics, how agents share information, coordinate reasoning, and avoid duplication of efforts.

Recent studies highlight the limitations of this evaluation and call for process-level metrics that assess the quality of intermediate reasoning steps~\cite{liu2023agentbench}. 
In practice, multi-agent systems often re-traverse the same inference paths or underutilize some agents entirely. 
Redundant messages can significantly inflate token usage, introducing a communication tax that increases both latency and computational cost~\cite{zhang2024cut}.

Existing evaluation metrics fail to expose these inefficiencies. 
Representing agent interactions as a directed graph, where nodes denote agents and edges represent message passing, provides a structured view of coordination patterns~\cite{zhang2024g}. 
This view enables the identification of redundant reasoning paths and inactive agents, offering actionable insight into the behavior of the system.
Furthermore, smaller open-source LLMs have demonstrated competitive performance with significantly reduced cost, in some cases up to 94\% lower than proprietary models~\cite{liu2023agentbench,zhang2024cut}. 
These trends motivate the need for evaluation methods that go beyond accuracy and help practitioners build more interpretable and efficient multi-agent systems.

Motivated by this gap, we present GEMMAS, a graph-based evaluation framework for analyzing multi-agent LLM reasoning processes.
GEMMAS encodes the entire reasoning trace as a directed acyclic graph (DAG), where each node represents an agent having (prompt, response) pair and each edge captures the flow of information between them.
We introduce two structure-aware metrics: 
(1) Information Diversity Score (IDS), which quantifies the semantic uniqueness of agent contributions, and 
(2) Unnecessary Path Ratio (UPR), which measures the fraction of reasoning steps that do not contribute new information. 
Together, these metrics evaluate collaboration efficiency beyond what is captured by task accuracy alone.

We apply GEMMAS to five mathematical reasoning benchmarks: GSM8K, AQuA, MultiArith, SVAMP, and MMLU using lightweight open-source LLMs. 
Our findings show that naïve agent pipelines suffer from high redundancy and low diversity, while configurations optimized for higher IDS and lower UPR improve both accuracy and token efficiency under fixed computational budgets.
GEMMAS thus surfaces hidden inefficiencies in multi-agent collaboration and offers practical design signals for building interpretable and cost-effective agent systems, an especially important concern in real-world industry deployments.

Our contributions are as follows:
\begin{itemize}
    \item We introduce GEMMAS, a graph-based evaluation framework for multi-agent LLM reasoning, and propose two novel structural metrics: Information Diversity Score (IDS) and Unnecessary Path Ratio (UPR), which together assess collaboration quality beyond task accuracy.
    \item We evaluate multiple small open-source LLMs, using GEMMAS to compare their collaborative behavior and identify efficiency trade-offs.
    \item We conduct systematic evaluations across five mathematical reasoning benchmarks, revealing substantial differences in collaboration quality among systems with similar final accuracies.
\end{itemize}

%% file: body/2_related_work.tex
\section{Related Work}
\label{sec:related_work}
\subsection{Agent-Level Diversity Metrics}
Standard metrics (e.g., BLEU, cosine similarity) capture surface-level variation~\cite{zhu2018texygen, li2015diversity} but ignore semantic roles and the graph structure critical for coordination~\cite{li2015diversity, park2024ensembling}.
Well-orchestrated lightweight models can match the performance of larger models in reasoning and planning~\cite{liu2023agentbench}, making efficient evaluation of agent-level diversity increasingly important.

Agent-level diversity in multi-agent systems requires specialized evaluation that considers how individual agents contribute unique perspectives to collaborative reasoning through their structural connectivity within the communication graph.
These limitations have motivated recent efforts to develop graph-aware metrics that consider both content and structural diversity in multi-agent reasoning.



\subsection{Evaluation in Multi-Agent Reasoning}
Evaluation of multi-agent reasoning remains in its early stages.
Traditional metrics such as final accuracy or task success rate often overlook the internal dynamics of agent collaboration.


To address limitations of conventional evaluations, graph-based methods have been proposed.
\textit{AgentPrune}~\cite{zhang2024cut} prunes low impact edges in the communication graph, while \textit{AgentDropout}~\cite{wang2025agentdropout} removes underperforming agents, both with the aim of reducing redundancy without compromising output quality. 
\textit{VillagerAgent}~\cite{villageragent} incorporates DAG based planning to assess workload balance and depth of reasoning.

These graph-based approaches highlight the growing emphasis on structural analysis to uncover inefficiencies and redundant communication patterns in collaborative reasoning among agents.

%



%% file: body/3_method.tex
\section{Evaluation Framework: GEMMAS}
\label{method}

In this section, we introduce \textbf{GEMMAS} (General Evaluation Metrics for Multi-Agent Systems), a comprehensive evaluation framework for graph-based multi-agent LLM systems. 
Unlike conventional approaches that focus solely on task outcomes, GEMMAS evaluates both the final results and the internal reasoning process of multi-agent collaboration.

\subsection{Task Definition and Problem Setup}
We consider the problem of multi-agent collaboration for mathematical reasoning tasks, where multiple language model agents interact through structured communication to solve problems jointly. 
Each multi-agent system (MAS) is modeled as a DAG, capturing the flow of information across agents throughout the reasoning process.

Formally, let $G = (V, E, F)$ denote the communication graph:
\begin{itemize}
    \item $V = \{v_1, v_2, ..., v_N\}$ is the set of $N$ agent nodes;
    \item $E \subseteq V \times V$ represents directed edges, where $(v_i, v_j) \in E$ indicates that the output of agent $v_i$ is available to agent $v_j$;
    \item $F = \{f_1, f_2, ..., f_N\}$ denotes the set of agent-specific reasoning functions, where $f_i$ defines the behavior or prompt processing logic of agent $v_i$.
\end{itemize}

To analyze both the communication structure and the temporal execution dynamics of the system, we maintain two adjacency matrices. 
The \textit{spatial adjacency matrix} $S \in \{0, 1\}^{N \times N}$ encodes direct communication links between agents, indicating which agents can exchange information. 
Complementarily, the \textit{temporal adjacency matrix} $T \in \{0, 1\}^{N \times N}$ captures the causal or time-ordered dependencies among agent outputs, allowing us to trace how intermediate reasoning steps influence one another across time.

GEMMAS evaluates multi-agent systems beyond final-task accuracy by analyzing the spatial and temporal structures of agent communication. 
This reveals inefficiencies such as redundant reasoning, low diversity, shallow chains, and idle agents, patterns not captured by conventional metrics.
While we report traditional baselines including accuracy (correct task completion rate) and token efficiency (prompt and completion token usage)~\cite{wang2025agentdropout}, GEMMAS provides process-level insights that quantify collaboration quality and structural resource utilization. 
By modeling both the topology and semantic content of communication graphs, it addresses a key limitation in current evaluation practices for multi-agent LLM systems.


\subsection{DAG-specific Metrics}

\textbf{Information Diversity Score (IDS).} 
This metric quantifies the heterogeneity of information generated by different agents by measuring the degree of similarity between their responses.
It addresses a fundamental question in collaborative systems: \textit{Do agents contribute unique perspectives, or do they merely repeat similar reasoning?}

Existing evaluation metrics are limited in multi-agent contexts, as they primarily capture surface-level correctness and ignore semantic intent or the structural role of each agent within the communication graph.
To address this gap, we combine syntactic analysis using TF-IDF~\cite{tf_idf} with semantic similarity computed via BERT embeddings~\cite{bert}, while incorporating structural context from the DAG.
To account for the topology of the agent graph, we weight each agent pair $(i, j)$ based on their spatial and temporal proximity within the DAG. 
The Information Diversity Score is defined as:
\begin{equation}
    IDS = \frac{\sum_{i, j} w_{ij} \cdot (1 - SS_{\text{total}}[i,j])}{\sum_{i,j} w_{ij}}
\end{equation}
\begin{equation}
    w_{ij} = \max(S_{ij}, S_{ji}) + \max(T_{ij}, T_{ji})
\end{equation}

Here, $SS_{\text{total}}[i,j]$ represents the average syntactic-semantic similarity between agents $i$ and $j$, computed as the cosine similarity of their TF-IDF and BERT representations, weighted equally with $\lambda_1 = \lambda_2 = 0.5$.
The weight $w_{ij}$ captures the relevance of agent pair $(i,j)$ based on their direct or indirect communication, using spatial adjacency matrix $S$ and temporal adjacency matrix $T$.

The complete algorithmic procedure for computing IDS, including similarity calculation and structure-aware weighting, is described in Algorithm~\autoref{alg:info_diversity}.
Sensitivity analysis for different weighting schemes is provided in Appendix~\ref{sec:appendix_ids}.

\vspace{0.5em}
\noindent\textbf{Unnecessary Path Ratio (UPR).} 
This metric assesses the structural efficiency of the MAS by identifying reasoning paths that provide negligible or redundant contributions to solving the task.
While IDS focuses on diversity, UPR addresses efficiency, quantifying the proportion of communication paths that fail to add meaningful information. 
It serves as an indicator of communication overhead and redundancy in agent interactions.

Formally, UPR is defined as:
\begin{equation}
    UPR = 1 - \frac{|\mathcal{P}_{\text{necessary}}|}{|\mathcal{P}_{\text{all}}|}
\end{equation}
where $\mathcal{P}_{\text{all}}$ represents the total number of reasoning paths in the spatial communication graph, and $\mathcal{P}_{\text{necessary}}$ includes only those paths that yield contribution scores above a predefined threshold.

A path is deemed necessary if it facilitates the production of correct or informative responses by downstream agents, based on a contribution function defined over message impact.
The detailed algorithmic pipeline, which includes path enumeration, contribution analysis, and threshold filtering, is outlined in Algorithm~\autoref{alg:upr}.

\begin{algorithm}[ht]
\small
\setstretch{1.25}
\caption{Information Diversity Score}
\label{alg:info_diversity}
\begin{algorithmic}[1]
\REQUIRE {Agent responses $O = \{o_1, \ldots, o_N\}$,\\%
    \hspace*{1.5em}Spatial adjacency matrix $S$,\\%
    \hspace*{1.5em}Temporal adjacency matrix $T$\\%
    }
\ENSURE Information diversity score $IDS \in [0, 1]$

\textit{/* Calculating syntactic-semantic similarity */}
\STATE Obtain syntactic features $\Phi \gets \text{TF-IDF(O)}$
\STATE Obtain semantic features $\Psi \gets \text{BERT(O)}$

\STATE $SS_{\text{syn}} \leftarrow \text{pairwise\_cosine}$($\Phi$)\hfill\COMMENT{syntactic similarity}
\STATE $SS_{\text{sem}} \leftarrow \text{pairwise\_cosine}$($\Psi$)\hfill\COMMENT{semantic similarity}
\STATE $SS_{\text{total}} \leftarrow \lambda_1 \cdot SS_{\text{syn}} + \lambda_2 \cdot SS_{\text{sem}}$

\textit{/* Calculating diversity score */}
\STATE Initialize weighted diversity $D_w \leftarrow 0$
\STATE Initialize DAG connection weights $W \leftarrow 0$
\FOR{$i = 1$ to $N - 1$}
    \FOR{$j = i+1$ to $N$}
        \STATE $w \gets \max(S_{ij}, S_{ji}) + \max(T_{ij}, T_{ji})$ \hfill\COMMENT{connection weights}

        \IF{$w > 0$} 
            \STATE $D_w \gets D_w + w \cdot (1- SS_{total}[i, j])$
            
            \STATE $W \gets W + w$
        \ENDIF
    \ENDFOR
\ENDFOR
\RETURN $D_w / W$
\end{algorithmic}
\end{algorithm}

\begin{algorithm}[ht]
\small
\setstretch{1.25}
\caption{Unnecessary Path Ratio}
\label{alg:upr}
\begin{algorithmic}[1]
    \REQUIRE {Spatial communication graph $G = (V, E_{\text{spatial}})$,\\%
    \hspace*{1.5em}Correct answer $\alpha$}
    \ENSURE Unnecessary path ratio $\text{UPR} \in [0, 1]$

    \textit{/* Path enumeration */}
    \STATE $\mathcal{P}_{all} \leftarrow \{p \mid p \text{ is a subpath in } G\}$
    \STATE $\mathcal{P}_{necessary} \leftarrow \emptyset$

    \textit{/* Path contribution assessment */}
    \FOR{each path $p \in \mathcal{P}_{all}$}
        \STATE Initialize correct count $c \leftarrow 0$
        \STATE Initialize total count $t \leftarrow 0$  
        
        \FOR{each agent $v \in p$}
            \STATE $a \leftarrow \text{ExtractAnswer}(\text{output}(v))$
            \IF{$a == \alpha$}
                \STATE $c \gets c + 1$
            \ENDIF
            \STATE $t \gets t + 1$
        \ENDFOR
        
        \STATE $\text{score} \gets \frac{c}{t}$ if $t > 0$ else $0$ \hfill\COMMENT{contribution score}
        \IF{$\text{score} \geq 0.5$}
            \STATE $\mathcal{P}_{\text{necessary}} \gets \mathcal{P}_{\text{necessary}} \cup \{p\}$
        \ENDIF
    \ENDFOR
    \RETURN $1 - |\mathcal{P}_{necessary}|/|\mathcal{P}_{all}|$ 
\end{algorithmic}
\end{algorithm}

\subsection{Evaluation Setup}
We systematically evaluate different MAS architectures that vary in their communication graph topologies.
Our evaluation encompasses both baseline and structurally optimized approaches to assess the effectiveness of DAG-based modifications.

\textbf{Baseline.} We employ a fully connected multi-agent system without structural optimization, where all agents can directly communicate with each other.
We refer to this baseline configuration as Vanilla-AD, as a reference point for measuring the effectiveness of communication graph modifications.

\subsubsection{Multi-Agent System Architecture} 
We adopt the agent-role configuration established in AgentPrune~\cite{zhang2024cut}, which defines a structured collaboration protocol among specialized agents to promote diversity and robustness in reasoning.

Our MAS setup involves four specialized agents with distinct roles. 
These are 
1) {AnalyzeAgent} focuses on problem decomposition and structured plan generation, 
2) {CodeWritingAgent} formulates code-based computational reasoning strategies, 
3) {MathSolverAgent} performs formal mathematical operations and symbolic solving and 
4) {AdversarialAgent} introduces plausible but intentionally incorrect solutions to stress-test robustness.

The collaborative process unfolds in three stages.
First, the input problem is distributed to all four agents concurrently.
Second, the agents engage in two rounds of communication based on the specified DAG topology, sharing intermediate reasoning traces from their specialized perspectives.
Finally, a FinalRefer agent aggregates these outputs to generate the final answer through collective reasoning.

\begin{figure}[t]
    \centering
    \includegraphics[width=\linewidth, keepaspectratio]{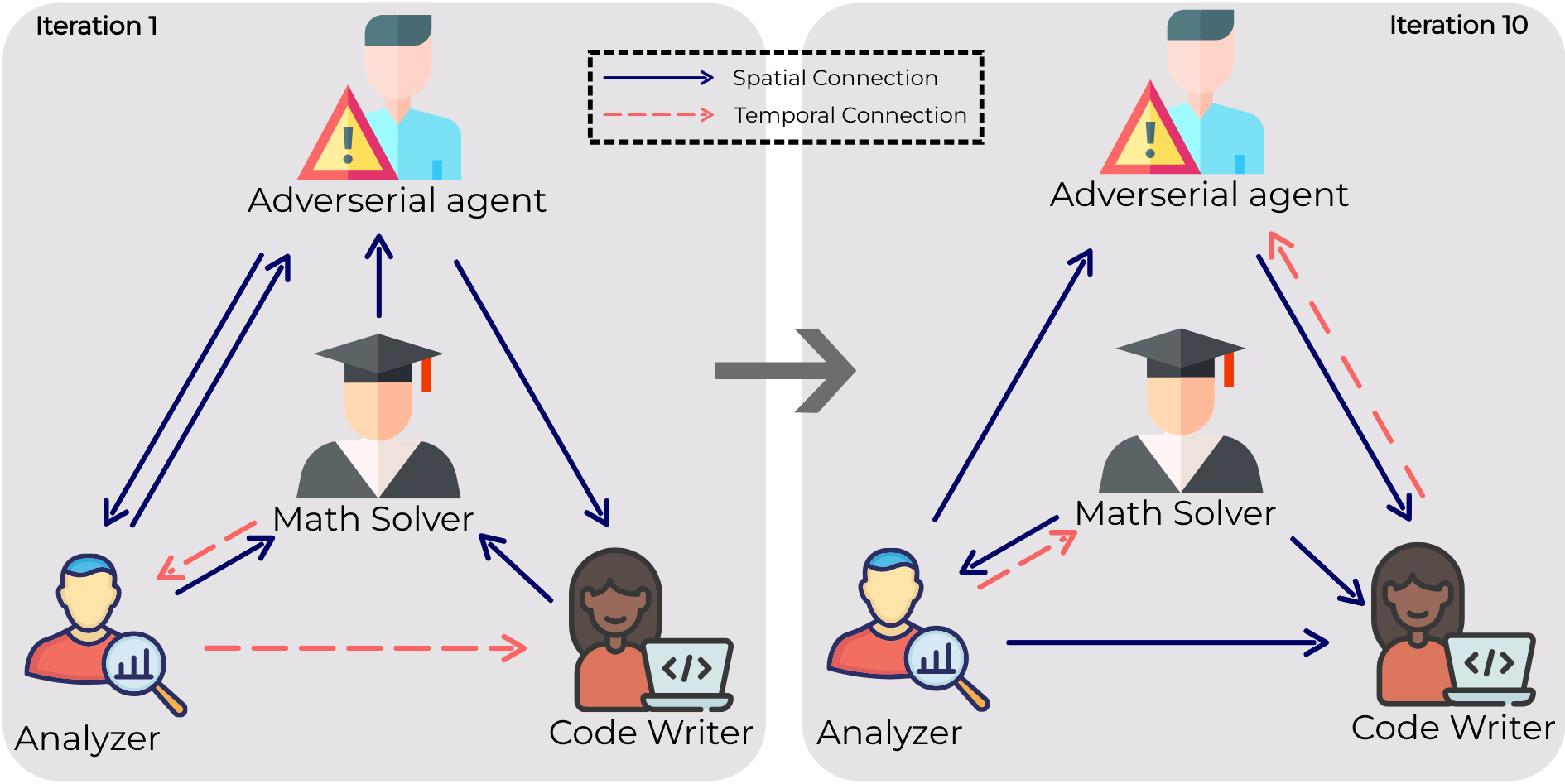}
    \caption{\textbf{Comparison of Multi-Agent DAG Structures Before and After Optimization.} 
    The figure shows the evolution of the DAG structure from the initial setup (Iteration 1) to the final optimized configuration (Iteration 10). 
    Solid blue lines denote spatial communication links, while dashed orange arrows indicate temporal dependencies.}
    \label{fig:fig2}
\end{figure}

\autoref{fig:fig2} illustrates the structural evolution of the DAG topology, comparing the initial structure at Iteration 1 with the optimized structure at Iteration 10 produced by the G-Designer method.
To analyze the effect of structural optimization, we implement and compare three state-of-the-art methods called 1) {AgentPrune} prunes communication links with low marginal impact, 2) {AgentDropout} dynamically removes underperforming agents and their associated links to reduce redundancy and 3) {G-Designer} learns optimal DAG topologies over multiple iterations, aiming to improve reasoning efficiency by minimizing communication overhead.

\subsubsection{Benchmarks and Language Models}
We evaluate DAG-based MAS architectures on five mathematical reasoning benchmarks: GSM8K~\cite{gsm8k}, which contains 1,000 grade-school math problems requiring multistep numeric reasoning; AQuA~\cite{aqua}, consisting of 254 algebraic word problems with multiple-choice answers; MMLU~\cite{mmlu}, where we select 748 questions from the mathematics subsets covering elementary to college-level topics; MultiArith~\cite{multiarith}, with 180 arithmetic word problems; and SVAMP~\cite{svamp}, comprising 300 elementary-level problems designed to test reasoning variation.
To examine how GEMMAS performs across different model scales, we use two small open-source instruction-tuned language models: Llama 3.1–8B-Instruct~\cite{llama3} and Qwen 2.5–7B-Instruct~\cite{yang2025qwen3}.

\subsubsection{Implementation Details}
All experiments are run under consistent hyperparameters: a learning rate of 0.1, a dropout rate of 0.1, 40 training examples, 10 sampling iterations, and two communication rounds per task. We set the generation temperature to zero for deterministic outputs. Unless stated otherwise, all other parameters follow the default configurations of the original MAS frameworks.

%% file: body/4_results_analysis.tex
\section{Results and Analysis}
\label{sec:results_analysis}

We apply \textsc{GEMMAS} to assess the structural quality and collaborative behavior of multi-agent systems (MAS). 
Tables~\ref{table:llama} and~\ref{table:qwen} present the results for both Llama3.1–8B-Instruct and Qwen2.5–7B-Instruct models across five reasoning benchmarks.

\begin{table}[ht]
  \centering
  \setlength{\tabcolsep}{3pt}
  \begin{adjustbox}{center}
  {\small
  \begin{tabular}{lccccc}
    \hline
    \textbf{Method } & \textbf{Accuracy\textcolor{red}{↑}} & \textbf{Ptok\textcolor{blue}{↓}} & \textbf{Ctok\textcolor{blue}{↓}} & \textbf{IDS\textcolor{red}{↑}} & \textbf{UPR\textcolor{blue}{↓}} \\
    \hline
    \rowcolor{gray!15}
    \multicolumn{6}{c}{GSM8K} \\
    \hline
    Vanilla - AD & 0.7961 & 10.18 & 3.16 & 0.33 & 0.39 \\
    AgentDropout & 0.6727 & \textbf{08.01} & \textbf{2.84} & \textbf{0.52} & 0.33 \\
    AgentPrune & 0.6688 & 12.68 & 4.11 & 0.33 & 0.32 \\
    G-Designer & \textbf{0.8391} & 11.09 & 3.43 & 0.32 & \textbf{0.14} \\

    \hline
    \rowcolor{gray!15}
    \multicolumn{6}{c}{AQuA} \\
    \hline
    Vanilla - AD & \textbf{0.6333} & \textbf{2.14} & \textbf{0.94} & \textbf{0.38} & 0.46 \\
    AgentDropout & 0.5833 & 2.37 & 1.07 & \textbf{0.38} & 0.47 \\
    AgentPrune & 0.5833 & 2.57 & 1.25 & 0.36 & \textbf{0.44} \\
    G-Designer & 0.5625 & 2.76 & 1.22 & 0.37 & 0.47 \\
    \hline
    \rowcolor{gray!15}
    \multicolumn{6}{c}{MultiArith} \\
    \hline
    Vanilla - AD & \textbf{0.9875} & 1.21 & 0.31 & 0.40 & 0.13 \\
    AgentDropout & 0.8688 & \textbf{0.99} & \textbf{0.26} & 0.40 & 0.24 \\
    AgentPrune & 0.8125 & 1.83 & 0.58 & \textbf{0.42} & 0.06 \\
    G-Designer & 0.9625 & 1.42 & 0.40 & 0.36 & \textbf{0.01} \\
    \hline
    \rowcolor{gray!15}
    \multicolumn{6}{c}{SVAMP} \\
    \hline
    Vanilla - AD & \textbf{0.8536} & \textbf{1.12} & \textbf{0.45} & 0.63 & \textbf{0.39} \\
    AgentDropout & 0.8000 & 1.35 & 0.58 & \textbf{0.66} & 0.46 \\
    AgentPrune & 0.8107 & 2.86 & 0.82 & 0.38 & 0.96 \\
    G-Designer & 0.8286 & 2.86 & 0.81 & 0.37 & 0.42 \\    
    \hline
    \rowcolor{gray!15}
    \multicolumn{6}{c}{MMLU} \\
    \hline
    Vanilla - AD & 0.5278 & 3.48 & 0.83 & 0.34 & 0.66 \\
    AgentDropout & 0.5389 & \textbf{2.47} & \textbf{0.68} & 0.34 & \textbf{0.62} \\
    AgentPrune & 0.5792 & 3.35 & 0.88 & 0.36 & 0.66 \\
    G-Designer & \textbf{0.7042} & 5.09 & 2.00 & \textbf{0.53} & 0.70 \\   
    \hline
  \end{tabular}
  }
  \end{adjustbox}
  \caption{Performance comparison of multi-agent systems on GSM8K, AQuA, MultiArith, SVAMP, and MMLU test dataset using Llama 3.1-8B-Instruct.}
  \label{table:llama}
\end{table}

\begin{table}[ht]
  \centering
  \setlength{\tabcolsep}{3pt}
  \begin{adjustbox}{center}
  {\small
  \begin{tabular}{lccccc}
    \hline
    \textbf{Method} & \textbf{Accuracy\textcolor{red}{↑}} & \textbf{Ptok}\textcolor{blue}{↓} & \textbf{Ctok}\textcolor{blue}{↓} & \textbf{IDS\textcolor{red}{↑}} & \textbf{UPR}\textcolor{blue}{↓} \\
    \hline
    \rowcolor{gray!15}
    \multicolumn{6}{c}{GSM8K} \\
    \hline
    Vanilla - AD & 0.8563 & 10.15 & 2.59 & 0.39 & 0.40 \\
    AgentDropout & 0.7797 & \textbf{06.99} & \textbf{1.58} & 0.40 & 0.41 \\
    AgentPrune & 0.7508 & 10.01 & 2.68 & 0.41 & 0.16 \\
    G-Designer & \textbf{0.8742} & 09.87 & 2.24 & \textbf{0.44} & \textbf{0.08} \\
    \hline
    \rowcolor{gray!15}
    \multicolumn{6}{c}{AQuA} \\
    \hline
    Vanilla - AD & \textbf{0.5958} & \textbf{1.82} & \textbf{0.73} & 0.38 & \textbf{0.31} \\ 
    AgentDropout & 0.5917 & 2.02 & 0.83 & 0.38 & 0.32 \\ 
    AgentPrune & 0.5417 & 1.98 & 0.77 & 0.49 & 0.34 \\
    G-Designer & 0.5042 & 2.06 & 0.80 & \textbf{0.52} & 0.36 \\
    \hline
    \rowcolor{gray!15}
    \multicolumn{6}{c}{MultiArith} \\
    \hline
    Vanilla - AD & 0.9938 & 1.20 & 0.24 & 0.43 & 0.16 \\
    AgentDropout & 0.9688 & \textbf{0.86} & \textbf{0.12} & 0.46 & 0.16 \\
    AgentPrune & 0.9938 & 1.45 & 0.38 & \textbf{0.57} & \textbf{0.00} \\
    G-Designer & \textbf{1.0000} & 1.12 & 0.21 &0.54 & \textbf{0.00} \\
    \hline
    \rowcolor{gray!15}
    \multicolumn{6}{c}{SVAMP} \\
    \hline
    Vanilla - AD & 0.8893 & \textbf{1.02} & 0.32 & 0.67 & 0.42 \\
    AgentDropout & \textbf{0.9071} & 1.05 & \textbf{0.29} & \textbf{0.69} & 0.36 \\
    AgentPrune & 0.8714 & 2.39 & 0.45 & 0.41 & 0.97 \\
    G-Designer & 0.9036 & 2.37 & 0.45 & 0.46 & \textbf{0.32} \\  
    \hline
    \rowcolor{gray!15}
    \multicolumn{6}{c}{MMLU} \\
    \hline
    Vanilla - AD & 0.7153 & 3.13 & 0.61 & 0.54 & 0.51 \\
    AgentDropout & 0.7181 & \textbf{2.17} & \textbf{0.47} & 0.63 & 0.53 \\
    AgentPrune & 0.7319 & 2.71 & 0.62 & 0.49 & \textbf{0.43} \\
    G-Designer & \textbf{0.7806} & 4.18 & 1.34 & \textbf{0.72} & 0.61 \\ 
    \hline
  \end{tabular}
  }
  \label{table:all-models-benchmarks}
  \end{adjustbox}
  \caption{Performance comparison of multi-agent systems on GSM8K, AQuA, MultiArith, SVAMP, and MMLU test dataset using Qwen2.5-7B-Instruct.}
  \label{table:qwen}
\end{table}

\subsection{Revealing Hidden Inefficiencies}
\label{sec:4.1}
Conventional evaluation metrics, such as final answer accuracy, fail to capture inefficiencies in multi-agent reasoning. 
For instance, on GSM8K with Qwen2.5-7B-Instruct, the Vanilla-AD configuration achieves 85.6\% accuracy, while G-Designer reaches 87.4\%. 
Although these results appear similar in terms of performance, GEMMAS reveals that G-Designer operates with significantly higher structural efficiency, recording a UPR of just 0.08 compared to 0.40 for Vanilla-AD, a fivefold improvement in redundant reasoning reduction.

Moreover, models with identical task accuracy may exhibit distinct internal collaboration patterns. 
On MultiArith with Qwen2.5-7B-Instruct, both Vanilla-AD and AgentPrune reach 99.4\% accuracy. 
However, AgentPrune demonstrates greater semantic diversity (IDS of 0.57 versus 0.43) and higher structural efficiency (UPR of 0.00 versus 0.16), highlighting that quality of reasoning is not reflected by accuracy alone.

\subsection{Identifying Optimal Configurations}

Building on these findings, we identify recurring trends in structural metrics that support actionable system design decisions. 
Systems that simultaneously exhibit high IDS and low UPR are particularly desirable, as they combine semantic diversity with efficient communication.

For example, AgentPrune on MultiArith consistently demonstrates this pattern. 
With Llama3.1–8B-Instruct, it achieves IDS 0.42 and UPR 0.06, while with Qwen2.5–7B-Instruct, it yields IDS 0.57 and UPR 0.00. 
In contrast, systems with low IDS and high UPR represent the least efficient configurations. 
On SVAMP, AgentPrune records IDS 0.38 and UPR 0.96 with Llama3.1–8B-Instruct, and IDS 0.41 and UPR 0.97 with Qwen2.5–7B-Instruct, signaling redundant or repetitive communication behavior.

From a performance efficiency perspective, some configurations manage to achieve strong accuracy with minimal communication overhead. 
For example, G-Designer on MMLU achieves 70.4\% precision with IDS 0.53 using Llama3.1–8B-Instruct, and 78.1\% precision with IDS 0.72 using Qwen2.5–7B-Instruct.

These results confirm that \textsc{GEMMAS} exposes structural trade-offs that traditional metrics cannot capture. 
It offers MAS designers a set of complementary signals to guide configuration choices based on specific goals, whether that is, maximizing computational efficiency (via low UPR), enhancing semantic richness (via high IDS), or achieving a balance across both dimensions depending on the deployment scenario.

%% file: body/5_conclusion.tex
\section{Conclusion}
\label{sec:conclusion}

We introduced \textsc{GEMMAS}, a graph-based evaluation framework for multi-agent language model systems that assesses collaboration quality beyond final-task accuracy. 
By modeling agent interactions as a DAG, GEMMAS defines two structural metrics—Information Diversity Score (IDS) and Unnecessary Path Ratio (UPR)—to capture semantic uniqueness and reasoning redundancy.
Experiments on five mathematical benchmarks reveal that systems with similar accuracy can vary significantly in internal collaboration patterns.
GEMMAS thus enables process-level diagnostics to guide the development of interpretable and efficient multi-agent systems.

\section*{Limitations}
While GEMMAS provides a structural lens to evaluate multi-agent LLM systems, it is currently limited to mathematical reasoning tasks and small open-source models. 
Extending this framework to broader domains, integrating runtime adaptivity, and coupling it with system-level metrics such as latency and memory footprint represent promising directions. 
Additionally, incorporating human-in-the-loop assessments and evaluating dynamic DAG topologies could further enhance the utility of GEMMAS for real-world deployment.

\section*{Acknowledgement}
This work was conducted as part of a collaboration between CARTE and LG Electronics, Toronto AI Lab.
We gratefully acknowledge the invaluable supervision and guidance of Homa Fashandi, Manasa Bharadwaj, and Kevin Ferreira throughout the course of this project.
This research was supported by the Institute of Information \& Communications Technology Planning \& Evaluation (IITP) grant funded by the Korean government (MSIT) [RS-2022-00143911, AI Excellence Global Innovative Leader Education Program].

%% file: body/appendix.tex
\appendix
\section{Appendix}
\label{sec:appendix}

\subsection{Information Diversity Score Analysis}
\label{sec:appendix_ids}
\begin{figure*}[htb]  
    \centering
    \includegraphics[width=\linewidth]{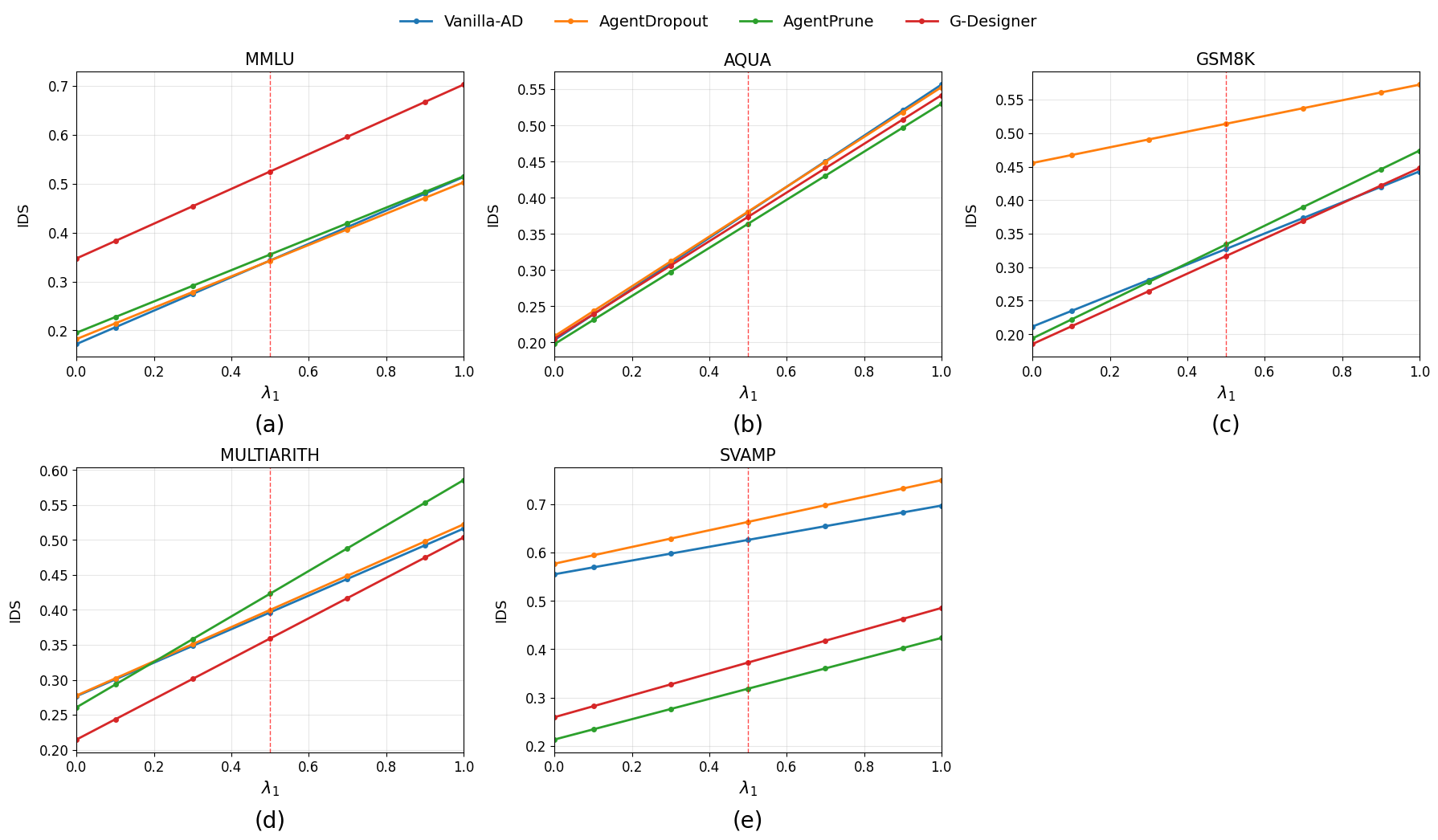}
    \caption{Sensitivity of Information Diversity Score (IDS) to syntactic-semantic weight balance across five benchmarks using Llama3.1-8B-Instruct. The x-axis shows $\lambda_1$ (syntactic weight), with vertical dashed line indicating $\lambda_1 = 0.5$ used in our main experiments.}
    \label{fig:ids_llama}
\end{figure*}

\begin{figure*}[htb]  
    \centering
    \includegraphics[width=\linewidth]{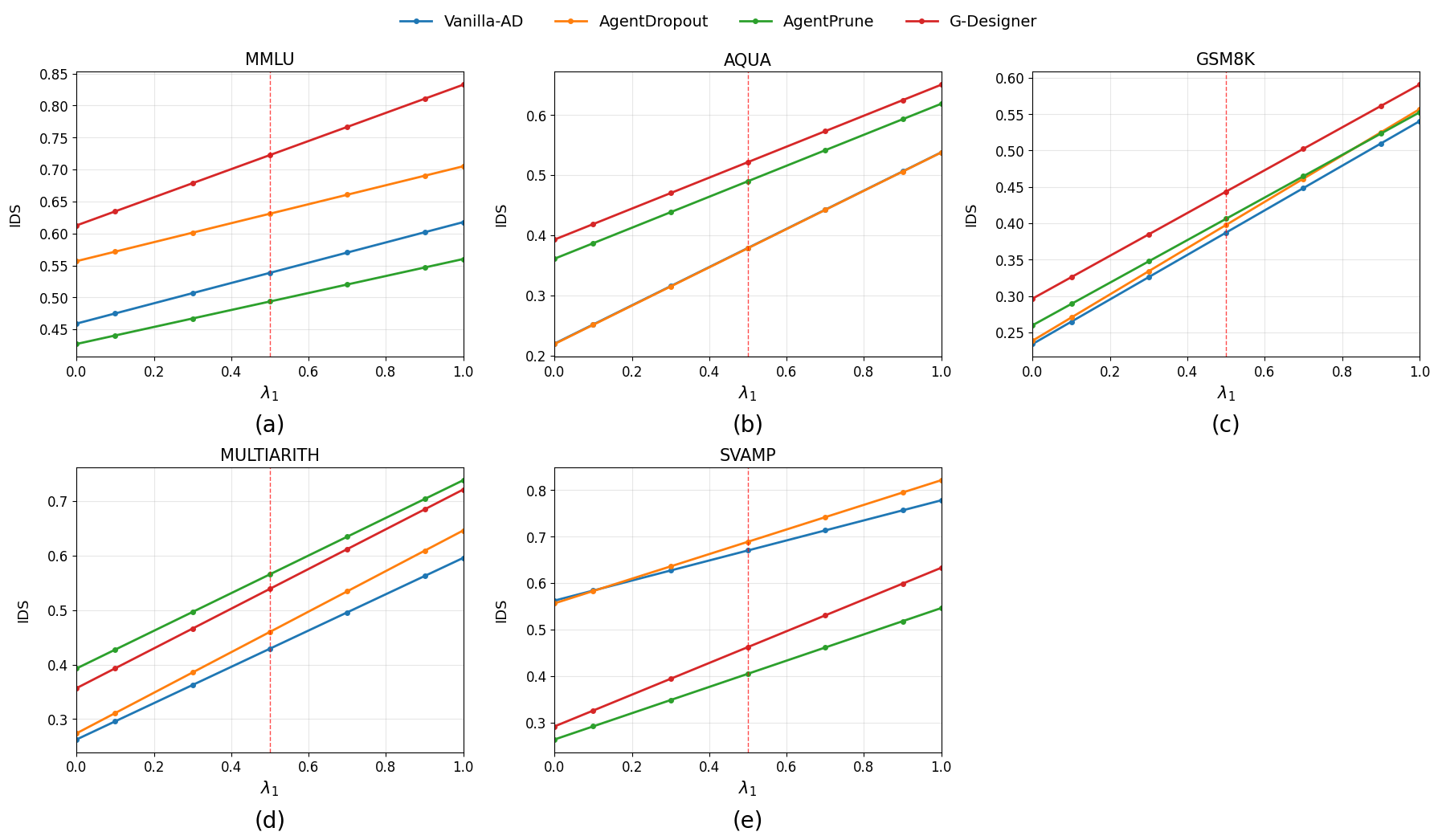}
    \caption{Sensitivity of Information Diversity Score (IDS) to syntactic-semantic weight balance across five benchmarks using Qwen2.5-7B-Instruct. The x-axis shows $\lambda_1$ (syntactic weight), with vertical dashed line indicating $\lambda_1 = 0.5$ used in our main experiments.}
    \label{fig:ids_qwen}
\end{figure*}

Figures~\ref{fig:ids_llama} and \ref{fig:ids_qwen} show how IDS values vary with different combinations of syntactic-semantic weights between benchmarks and models. 
The weight balance parameter $\lambda_1$ controls the contribution of syntactic characteristics (TF-IDF), while semantic features (BERT embeddings) are weighted as $\lambda_2=1-\lambda_1$.

As $\lambda_1$ increases toward 1.0, emphasizing syntactic similarity, IDS values generally increase between models and benchmarks. 
This suggests that syntactic diversity (e.g., different vocabulary usage, sentence structures) tends to be more pronounced than semantic diversity in multi-agent communications.

Different multi-agent systems exhibit varying sensitivity to weight balance.
For example, on SVAMP with both models, Vanilla-AD and AgentDropout show relatively higher IDS values compared to AgentPrune and G-Designer, indicating more diverse communication patterns across different syntactic-semantic weight settings.

We selected equal weights to balance the syntactic and semantic contribution without favoring either aspect of the similarity measurement.
This balanced approach provides a neutral baseline for comparing multi-agent systems across different collaboration patterns.